\documentclass[letterpaper, 11 pt]{article}  % % Comment this line out
\usepackage[left=2.6cm, right=2.6cm, top=2cm, bottom=2cm]{geometry}                     

\usepackage{lipsum}

\newcommand\blfootnote[1]{%
  \begingroup
  \renewcommand\thefootnote{}\footnote{#1}%
  \addtocounter{footnote}{-1}%
  \endgroup
}

\usepackage{subcaption}
\usepackage{mwe}
\usepackage{algcompatible}
\usepackage[linesnumbered,ruled,vlined]{algorithm2e}
\usepackage{balance} %even columns on last page

\usepackage{xcolor,calc}
\usepackage{graphics}
\usepackage{hyperref}
\usepackage{epstopdf}
\usepackage{graphicx}
\usepackage{amsmath} % assumes amsmath package installed
\usepackage{amssymb}  % assumes amsmath package installed
\usepackage{amsthm}
\usepackage[noadjust]{cite}
\usepackage{color}
\usepackage{enumerate}
\usepackage{multirow}
\usepackage{mathtools}
\usepackage{booktabs}
\usepackage{epstopdf}
\usepackage[normalem]{ulem}%%%%%%%%%

\usepackage{url}
\usepackage{bm}
\usepackage{caption}

\usepackage{booktabs}

\newtheorem{remark}{Remark}

\makeatletter
\newsavebox\myboxA
\newsavebox\myboxB
\newlength\mylenA

\newcommand*\xoverline[2][0.75]{%
    \sbox{\myboxA}{$\m@th#2$}%
    \setbox\myboxB\null% Phantom box
    \ht\myboxB=\ht\myboxA%
    \dp\myboxB=\dp\myboxA%
    \wd\myboxB=#1\wd\myboxA% Scale phantom
    \sbox\myboxB{$\m@th\overline{\copy\myboxB}$}%  Overlined phantom
    \setlength\mylenA{\the\wd\myboxA}%   calc width diff
    \addtolength\mylenA{-\the\wd\myboxB}%
    \ifdim\wd\myboxB<\wd\myboxA%
       \rlap{\hskip 0.5\mylenA\usebox\myboxB}{\usebox\myboxA}%
    \else
        \hskip -0.5\mylenA\rlap{\usebox\myboxA}{\hskip 0.5\mylenA\usebox\myboxB}%
    \fi}
\makeatother

\usepackage{tabularx}
\usepackage{booktabs}

\newtheorem{definition}{Definition}
\newtheorem{assumption}{Assumption}

\makeatletter
\let\NAT@parse\undefined
\makeatother

\begin{document}

\title{Learning  Environment  Constraints  in  Collaborative  Robotics: \\ A  Decentralized  Leader-Follower Approach}
% \title{Robust MPC for LPV Systems}

\author{Monimoy Bujarbaruah, Yvonne R. St{\"u}rz, Conrad Holda, \\
Karl H. Johansson, and Francesco Borrelli\blfootnote{Emails:\{monimoyb, y.stuerz, conradholda, fborrelli\}@berkeley.edu, kallej@kth.se.}
}

\maketitle

%%%%%%%%%%%%%%%%%%%%%%%%%%%%%%%%%%%%%%%%%%%%%%%%%%%%
\begin{abstract}
    In this paper, we propose a leader-follower hierarchical strategy for two robots collaboratively transporting an object in a partially known environment with obstacles. Both robots sense the local surrounding environment and react to obstacles in their proximity. We consider no explicit communication, so the local environment information and the control actions  are not shared between  the robots.
    % , and instead assume only implicit communication through force feedback. 
    At any given time step, the leader solves a model predictive control (MPC) problem with its known set of obstacles and plans a feasible trajectory to complete the task. The follower estimates the inputs of the leader and uses a policy to assist the leader while reacting  to obstacles in its proximity. The leader  infers obstacles in the follower's vicinity by 
    using the difference between the predicted and the real-time estimated follower control action.
% The set of known obstacle avoidance constraints for the leader's MPC problem is then updated at the next step. 
A method to switch the leader-follower roles is used to  improve the control performance in tight environments. The efficacy of our approach is demonstrated with detailed comparisons to two alternative strategies, where it achieves the highest success rate, while completing the task fastest. See \href{https://www.dropbox.com/s/hexadigqkvspaeh/IROS_Video.mp4?dl=0}{\textit{www.dropbox.com/s/hexadigqkvspaeh/IROS\_Video.mp4?dl=0}} for a descriptive video of the algorithm.    
\end{abstract}
%%%%%%%%%%%%%%%%%%%%%%%%%%%%%%%%%%%%%%%%%%%%%%%%%%%
\section{Introduction}
Multi-robot systems have a high potential to collaboratively accomplish complex tasks, such as, for example transporting large or heavy work pieces \cite{feng2020overview, khatib1987unified, khatib1996coordination, rus1995moving, song2002potential, wang2002object,sugie1995placing}. For known repetitive tasks in structured environments, collaborative manipulation problems in industrial applications are mainly solved in a centralized way, relying on precise feedforward computations. As solving a centralized control synthesis problem for such tasks can become computationally cumbersome,  distributed and decentralized control strategies have been proposed \cite{donald2000distributed,li2008robust,franchi2018distributed,marino2017distributed,habibi2015distributed,dai2015leaderless,sturz,zhu}. However, communication delays remain as a bottleneck, i.e., iterative communication, such as the ones required for consensus type algorithms, might converge too slowly for such robotics applications. To resolve this, the use of implicit communication, such as force and torque measurements or estimates on rigid bodies, has been proposed in the literature, e.g., in \cite{losey2020learning, swag, wang2016force, stilwell1993toward, gross2006transport, aiyama1999cooperative, kosuge1996decentralized, takeda2004load, gross2004group,marino2018two}. 

Obstacle avoidance in collaborative robotics has primarily considered known obstacles and solving a centralized problem with explicit communication \cite{alonso2015local,li2019dynamical, brock2002task, desai1999motion,bharatheesha2017dynamic,shorinwa2020scalable}.
Issues arise when the robots rely only on local controllers in unstructured and unknown or partially known environments, primarily
% . This is challenging for a cooperative manipulation task
because of the tight dynamical couplings. 
A particular challenge is associated with inferring unknown obstacles using implicit communications when the robots have only partial knowledge of their environment, i.e. they use local sensors with limited field of view. 

We consider the task of two robots collaboratively transporting an object, constraining the robots' inputs to comply with the object's physical constraints. We  consider  no  explicit communication, so the local environment information and the  control  actions  are  not  shared  between  the  robots. We solve the control design problem by using a leader-follower strategy with the leader using a predictive control and the follower using a simple controller, known to the leader. 
With this schema, the leader can solve the collaborative transportation task, with the help of the follower, while building a map of its unknown obstacles.
Such a map is obtained by  estimating the  follower’s  inputs to infer missing local information about the  environment sensed by the follower. This extends the work of \cite{petrov1981control,dumonteil2015reactive} to two-robot problems. Our key contributions can be summarized as follows:
\begin{enumerate}
    \item We propose a leader-follower strategy for two robots collaboratively transporting an object in a partially known environment with obstacles. The leader solves an MPC problem based on its known set of obstacles and plans a trajectory to reach the target position, while avoiding collisions for the whole system (i.e., the two robots combined with the object to be transported). 
    \item We present  a simple control policy for the follower that is reactive to obstacles detected by the follower (and possibly undetected by the leader). This follower control policy  is designed so that it allows the leader to infer the position of  obstacles not directly sensed. %The leader accordingly updates its known set of obstacles before the next planning step. 
    \item Motivated by \cite{losey2020learning}, we introduce a strategy for allowing leader-follower role switches during the task. We present a detailed numerical example of two point robots transporting a rigid rod in an initially unknown environment. On this example our proposed approach allows the leader's MPC controller to learn the undetected obstacles and successfully complete the task, with the leader-follower roles appropriately switched.  
    % demonstrate that with such a switching strategy, the joint system avoids obstacles in tight spaces, which it otherwise fails to do.
    
    % \item We demonstrate the effectiveness of the proposed approach through an initially unknown tightly constrained environment. We show that the obstacles are successfully avoided when the leader-follower roles are appropriately switched. 
\end{enumerate}
% We have limited ourselves to two robots. 
Control design with three or more robots is not addressed in this work. In order to estimate the inputs of the other robot, we assume each robot can estimate the states of the joint system, i.e., the two robots with the object to be transported. For the considered example of two point robots transporting a rigid rod, this estimation is done with measurements of robots' own positions and the rod orientation. For more complicated systems, similar estimates may be obtained using additional sensors. We do not present this in this paper. 
\section{Problem Formulation}
In this section, we formulate the collaborative obstacle avoidance problem with the leader-follower control architecture. Such a  leader-follower hierarchy is common in control design \cite{bemporad2011decentralized, swag, franze2018distributed, losey2020learning}. We limit ourselves to the case of only one follower. We refer to the two robots with the object to be transported as the \emph{joint system}.
\subsection{Environment Constraints}
Let the environment be defined by the set $\mathcal{X}$. %For this work w
We assume obstacles are static, although the proposed framework can be extended to dynamic obstacles. Let the set of obstacles be denoted by $\mathcal{O}$. Therefore, the safe set for the joint system is given by $\mathcal{S} = \mathcal{X} \setminus \mathcal{O}$. At the beginning of the task, we assume that the robots do not have any prior information about the environment. During the control task, the robots detect obstacles and store their positions. At any time step $t$, let the set of obstacle constraints known to the leader and the follower (detected at $t$ and stored until $t$) be denoted by $\mathcal{C}_{l,t}$ and $\mathcal{C}_{f,t}$, respectively. We denote:
\begin{align*}
    &\mathcal{C}_{l,t} \cup \mathcal{C}_{f,t} = \mathcal{O}_t,~\textnormal{with}~ \mathcal{O}_{t-T_s} \subseteq \mathcal{O}_t,~\forall t \leq T, 
\end{align*}
where $T_s$ is the sampling time of both the leader and the follower robot controllers (defined next in Section~\ref{ssec:sys_modeling}), and  $T \gg T_s$ is the task duration.
\subsection{System Modeling}\label{ssec:sys_modeling}
\begin{figure}[h!]
	\centering
	\includegraphics[width=0.7\columnwidth]{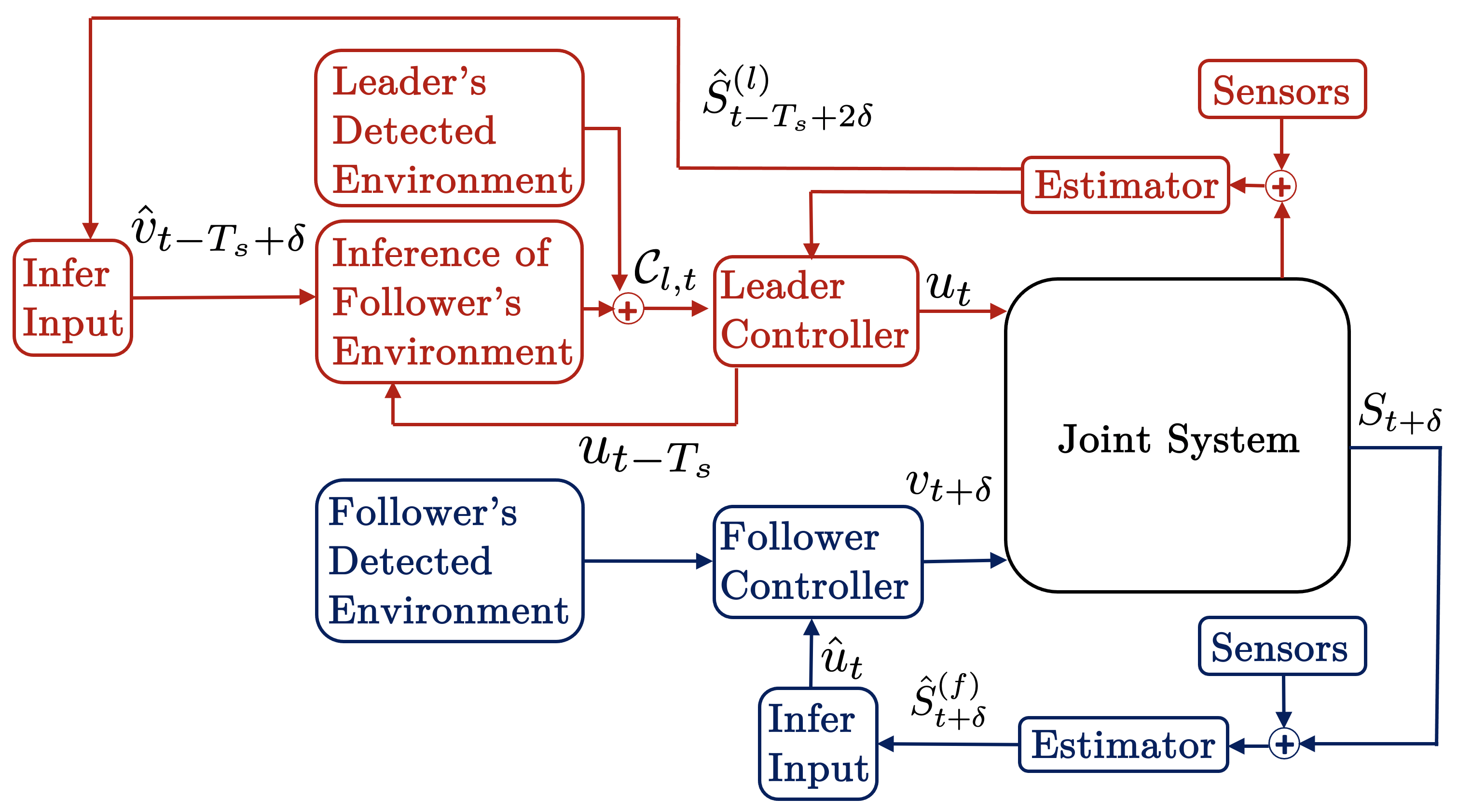}
	\caption{Block diagram of the joint system with leader follower controllers. 
	}
	\label{fig:blk}    
\end{figure}

We consider that the leader and the follower robots transport the same object as they move. The state space equation of the joint system is of the form:
\begin{align}\label{eq:mod_gen}
    S_{t+T_s} = f(S_t, u_t, v_t),
\end{align}
where $S_t \in \mathbb{R}^d$ is the joint system state, $u_t \in \mathbb{R}^m$ is the input of the leader and $v_t \in \mathbb{R}^p$ is the input of the follower at time step $t$, and $f(\cdot, \cdot, \cdot)$ is any nonlinear map. 
\begin{remark}\label{rem:rem_Rt}
In general the states $S_t$ contain the positions and velocities of the center of masses of the leader, the follower and the object being transported. 
\end{remark}
% We discretize system \eqref{eq:mod_gen} using one step forward Euler integration with step size $T_s$ to obtain, 
% \begin{align}\label{eq:sys_model}
%     S_{t+1} & = S_t + T_s f_c(S(tT_s), u(tT_s), v(tT_s)), \nonumber \\
%     & = f(S_t, u_t, v_t),
% \end{align}
% with $S_0 = S(0)$, $u_t = u(tT_s)$ and $v_t = v(tT_s)$ for all discrete time steps $t = \{0,1,\dots\}$. Note that between each two sampling steps we utilize a zero order hold strategy for applying the inputs, i.e., 
% \begin{subequations}\label{eq:zoh}
% \begin{align}
%     & u(\tau) = u_t,~\forall \tau \in [tT_s, (t+1)T_s),  \\
%     & v(\tau) = v_t,~\forall \tau \in [tT_s, (t+1)T_s).  
% \end{align}
% \end{subequations}
% \textbf{MONIMOY LET'S TALK ABOUT THIS DIAGRAM-- do not like that blue line into red.. then it is no longer decentralized.. or is just know knowledge of the control law?} 
A block diagram of the joint system is shown in Fig.~\ref{fig:blk}, where the red and the blue parts indicate the operations carried out by the leader and the follower, respectively.  
We consider the case where the leader does not have full information of all the detected obstacles in $\mathcal{O}_t$, i.e., $\mathcal{C}_{l,t} \subset \mathcal{O}_t$. We further consider that no explicit communication between the leader and the follower is available. Similar to \cite{losey2020learning, swag, wang2016force, stilwell1993toward, gross2006transport, aiyama1999cooperative, kosuge1996decentralized, takeda2004load, gross2004group}, we enable both the agents to infer each other's inputs as ``implicit" communication. We first make the following assumption. 
\begin{assumption}\label{assump:stick_info}
The leader and the follower can estimate the joint system states $S_t$ at all time steps.
% The leader can estimate the relevant system states that are needed for the obstacle avoidance task.
\end{assumption}
We introduce the following notation: let $(\cdot)^{(j)}_{t}$ denote the value of the quantity $(\cdot)_{t}$ as inferred by robot $j \in \{l,f\}$. The leader and the follower's estimates of $S_t$ are thus denoted by $\hat{S}^{(l)}_t$ and $\hat{S}^{(f)}_t$, respectively. We denote the coordinates of the center of mass of the follower by $R_t = [X_{f,t}, Y_{f,t}]^\top$, and the leader/follower estimates $\hat{R}^{(l/f)}_t = [\hat{X}^{(l/f)}_{f,t}, \hat{Y}^{(l/f)}_{f,t}]^\top$. Often such states are already included in $\hat{S}^{(l/f)}_t$, as pointed out in Remark~\ref{rem:rem_Rt}. If they are not a part of $\hat{S}^{(l/f)}_t$, they need to be estimated as well in our control approach. 

\vspace{3pt}
% \textbf{the timing here is messy!}
\noindent \textbf{Method Outline:} At time step $t$, the leader uses $\hat{S}^{(l)}_t$ to  compute the control action $u_t$ for the joint system to avoid its known set of obstacles $\mathcal{C}_{l,t}$. As there is no explicit communication, the follower infers the leader's inputs $u_t$ via its state estimates, inducing a delay in the application of its inputs. That is, at time step $t+\delta$ (with a $\delta \ll T_s$), the follower uses  $\hat{S}^{(f)}_{t+\delta}$ and $\hat{R}^{(f)}_{t+\delta}$ to infer $\hat{u}_t$ . The follower also uses $\hat{R}^{(f)}_{t+\delta}$ to build a map of its detected obstacles, and computes $v_{t+\delta}$ as a function of $u_t$ and these obstacles. During the inference time between $t$ and $t+\delta$ the follower keeps applying the previous input $v_{(t-T_s)+\delta}$. The leader then infers the follower's inputs $v_{t+\delta}$ via its state estimates to learn additional obstacles. That is, at time step $(t+2\delta)$\footnote{the leader's inference can be made at time step $(t+\delta+\mu)$ with $\delta<\mu \leq T_s$. We use $\delta \ll T_s$ and $\mu=\delta$ only to simplify notations. See Table~\ref{tab1}.}, the leader uses $\hat{S}^{(l)}_{t+2\delta}$ to estimate  $\hat{v}_{t+\delta}$, based on which it learns the position of any additional obstacles in the follower's proximity at $t+\delta$ using $\hat{R}^{(l)}_{t+\delta}$. The leader then computes updated $\mathcal{C}_{l,t+T_s}$. We detail the algorithm in Section~\ref{sec:control_synth}. In the next sections, we present the controller synthesis. We discuss the effect of the  time delay $\delta$ in details in Section~\ref{ssec:fol_inp_infer}-\ref{ssec:learn_obs}, when we distinguish between the leader and the follower applied inputs.

\begin{remark}
We consider that the leader and follower robots have synchronized clocks. A short  discussion of non synchronized clocks is presented in Section~\ref{sec:alg}.
\end{remark}

% \textbf{MY SUGGESTION: IN THE METHOD OUTLINE, USE THE RIGHT TIMING , THEN YOU CAN SAY THAT YOU ASSUME DELTA IS VERY SMALL AND SET ALL TO ZEROS FOR NOTAtION SIMPLICITY, AND THE DISCUS DELTA IN DETAILS IN SECTION XYZ}

%%
% Assumption~\ref{assump:stick_info} can be relaxed and noisy output feedback can be incorporated, with the design of appropriate state observers, such as a particle filter or an extended Kalman filter. For now, we leave this as a future extension of our research. 

\begin{figure}[h!]
	\centering
	\includegraphics[width=0.55\columnwidth]{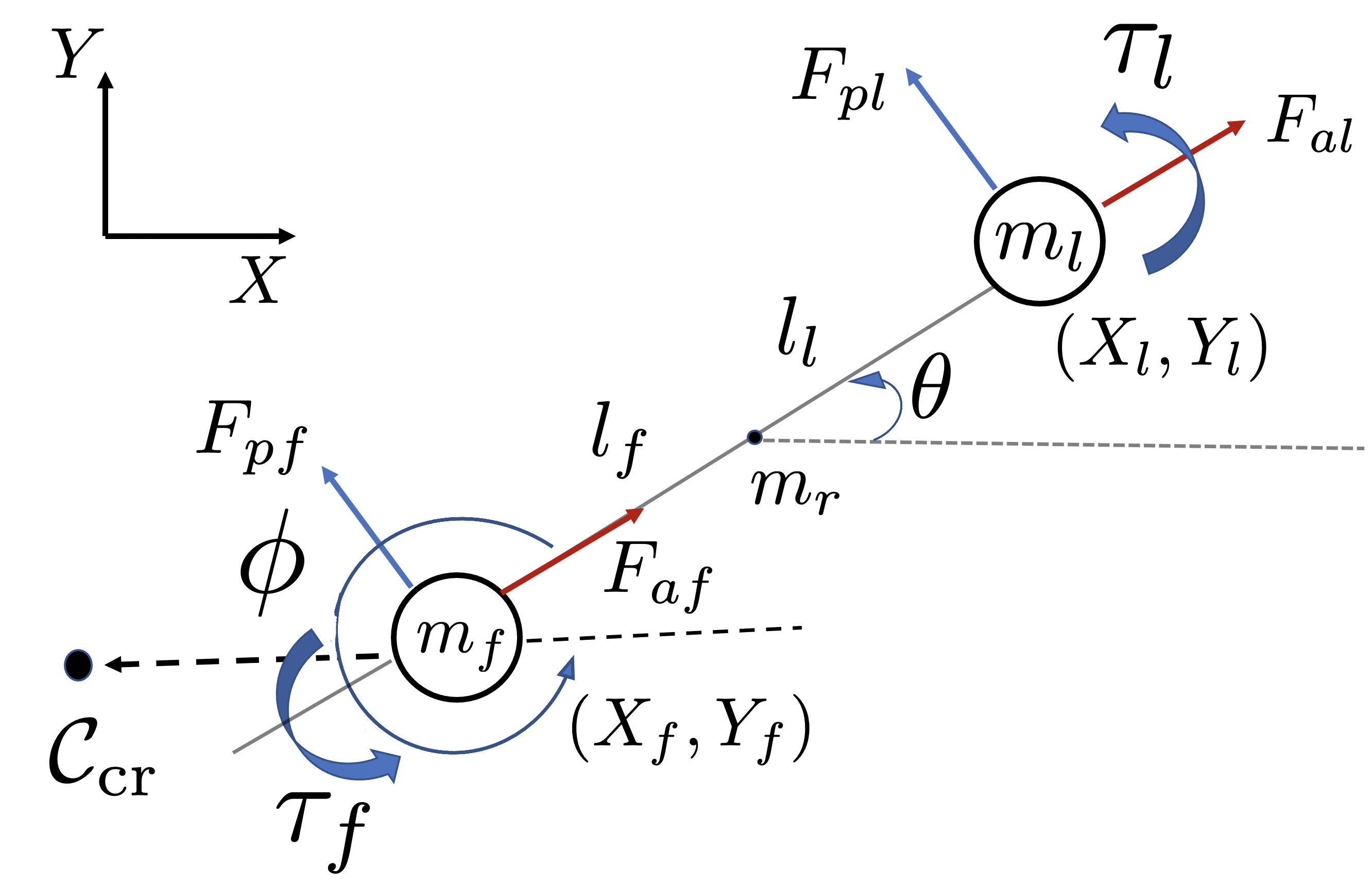}
	\caption{Model of the joint system. The leader and the follower are point masses connected by the rigid rod. The follower reacts to a \emph{critical obstacle} $\mathcal{C}_{\mathrm{cr}}$, as defined in Section~\ref{ssec:fol_pol_ssec}.}
	\label{fig:sysmodel}    
\end{figure}

For clarity of presentation in this paper, we present a specific case of model \eqref{eq:mod_gen}. Specifically, we model both the leader and the follower as point robots $m_l$ and $m_f$, with global coordinates $X_l, Y_l$ and $X_f, Y_f$, respectively, transporting a rigid rod, as shown in Fig.~\ref{fig:sysmodel}. 
% The connecting rod of length $(l_l+l_f)$ has a moment of inertia $J$, with a mass negligible compared to $m_l$ and $m_f$.
The connecting rod of length $(l_l+l_f)$ has a mass $m_r$. Assuming the rod is of uniform density, the rod has an inertia, $J_r$, of $\frac{1}{12}m_r (l_l+l_f)^2$.
The total mass of the joint system is therefore $m=m_r+m_l+m_f$. The total moment of inertia of the joint system is $J = J_r + m_r(\frac{l_l-l_f}{2})^2 + m_l l_l^2 + m_f l_f^2$.
The leader's inputs on the rod are the axial force $F_{al}$, perpendicular force $F_{pl}$, and the torque $\tau_l$. The corresponding follower's  inputs are $F_{af}$, $F_{pf}$ and $\tau_f$.  Denoting $\mathcal{T} = \frac{(-F_{pf}l_f + F_{pl}l_l + \tau_l+\tau_f)}{J}$ and define:
% \begin{equation}\label{eq:t_ex}
% \begin{aligned}
%     q_1 & = -(l_l \sin \theta \frac{(-F_{pf}l_f + F_{p1l}l_l)}{J} + l_l \cos \theta \dot{\theta}^2) + \cdots \\ 
%     & ~ + \frac{1}{(m_l+m_f)} (\cos \theta (F_{al} + F_{af}) - \sin \theta (F_{pl} + F_{pf})), \\
%     q_2 & = -(-l_l \cos \theta \frac{(-F_{pf}l_f + F_{pl}l_l)}{J} + l_l \sin \theta \dot{\theta}^2) + \cdots \\
%     & ~ + \frac{1}{(m_l+m_f)} (\sin \theta (F_{al} + F_{af})) + \cos \theta (F_{pl} + F_{pf})).
% \end{aligned}
% \end{equation}
\begin{equation}\label{eq:t_ex}
\begin{aligned}
    q_1 & = -(l_l \sin \theta \mathcal{T} + l_l \cos \theta \dot{\theta}^2) + \frac{1}{m} (\cos \theta (F_{al} + F_{af}) - \sin \theta (F_{pl} + F_{pf})), \\
    q_2 & = (l_l \cos \theta \mathcal{T} - l_l \sin \theta \dot{\theta}^2) + \frac{1}{m} (\sin \theta (F_{al} + F_{af})) + \cos \theta (F_{pl} + F_{pf})).
\end{aligned}
\end{equation}
Due to the rigid coupling with the rod, the leader's position, and translational and angular velocity states are sufficient to define the evolution of the joint system. Accordingly, using \eqref{eq:t_ex}, the state-space equation for the joint system is: 
\begin{align}\label{eq:mod_con}
    \dot{S}(t) & = f_c(S(t), u(t), v(t)) = \begin{bmatrix} \dot{X}_l & q_1 & \dot{Y}_l & q_2 & \dot{\theta} & \mathcal{T} \end{bmatrix}^\top,
\end{align} 
with $S(t) = [s_1, s_2, s_3, s_4, s_5, s_6]^\top$, 
$u(t) = [F_{al}, F_{pl}, \tau_l]^\top$, and $v(t) = [F_{af}, F_{pf}, \tau_f]^\top$ at time $t$, where the states are representative of variables given by:
\begin{align*}
    s_1 = X_l, s_2 = \dot{X}_l, s_3 = Y_l, s_4 = \dot{Y}_l, s_5 = \theta, s_6 = \dot{\theta}.
\end{align*}
We discretize \eqref{eq:mod_con} with the sampling time of $T_s$ for both the leader and the follower to obtain its discrete time version similar to \eqref{eq:mod_gen}.
% Between each two sampling steps we utilize a zero order hold strategy, i.e., 
% \begin{subequations}\label{eq:zoh}
% \begin{align}
%     & u(\tau) = u_t,~\forall \tau \in [tT_s, (t+1)T_s),  \\
%     & v(\tau) = v_t,~\forall \tau \in [tT_s, (t+1)T_s),  
% \end{align}
% \end{subequations}
% for all discrete time steps $t = \{0,1,\dots, T\}$. 
Furthermore, for this specific model \eqref{eq:mod_con}, Assumption~\ref{assump:stick_info} can be stated as: both the leader and the follower can estimate the leader's position, velocity, as well as the angular speed and orientation of the rod at all times. For simplicity of presentation, we consider that the robots measure their positions, velocities, and the rod's angle and angular speed. Accordingly, estimators $\hat{S}^{(l)}_t$ and $\hat{S}^{(f)}_{t}$ are: 
\begin{subequations}\label{eq:estimates_us}
\begin{align}
        & \hat{S}^{(l)}_t = \begin{bmatrix} {X}_{l,t} & \dot{{X}}_{l,t} & {Y}_{l,t} & \dot{{Y}}_{l,t} & \theta_t & \dot{\theta}_t \end{bmatrix}^\top, \label{eq:lead_states}\\ 
        & \hat{S}^{(f)}_{t} = 
        \begin{bmatrix} \hat{X}^{(f)}_{l,t} \\ \dot{\hat{X}}^{(f)}_{l,t} \\ \hat{Y}^{(f)}_{l,t} \\ \dot{\hat{Y}}^{(f)}_{l,t} \\ \theta_t \\ \dot{\theta}_t \end{bmatrix} = \begin{bmatrix}  X_{f,t} + (l_l + l_f) \cos \theta_t \\ \dot{X}_{f,t} - (l_l + l_f) \sin \theta_t \dot{\theta}_t \\ Y_{f,t} + (l_l + l_f) \sin \theta_t \\ \dot{Y}_{f,t} + (l_l + l_f) \cos \theta_t \dot{\theta}_t \\ \theta_t \\ \dot{\theta}_t \end{bmatrix} \label{eq:fol_states}.
\end{align}
\end{subequations}
In the absence of perfect position, velocity and rod orientation measurements, one can design appropriate state observers, such as a particle filter or an extended Kalman filter to obtain their estimates, if Assumption~\ref{assump:stick_info} holds. 
% We leave this as a future extension of our research. 

%%
% \begin{remark}\label{rem:param_est}
% Note, we have assumed that all the masses and the moment of inertia values are known. If these are unknown, then they can be estimated offline and such estimates can be used for control design. 
% \end{remark}
%%
\subsection{Input Constraints}
We consider constraints on the inputs of both the leader and the follower, which are given by $u_t \in \mathcal{U}$ and $v_t \in \mathcal{V}$ for all $t \geq 0$. For our specific example in this paper, with $\bar{F}_a, \bar{F}_p, \bar{\tau} \in \mathbb{R}_+$, we consider the same box constraints:
\begin{align}\label{eq:inp_con}
    \mathcal{U} = \mathcal{V} := \{w: -\begin{bmatrix} \bar{F}_a & \bar{F}_p & \bar{\tau} \end{bmatrix}^\top \leq w \leq \begin{bmatrix} \bar{F}_a & \bar{F}_p & \bar{\tau} \end{bmatrix}^\top \}.
\end{align}
%%
% \subsection{Environment Constraints}
% Let the environment be defined by the set $\mathcal{X}$. For this work we assume obstacles are static, although the proposed framework can be extended in presence of dynamic obstacles. Let the set of obstacles be denoted by $\mathcal{O}$. Therefore, the \emph{safe set} for the joint system is given by: 
% \begin{align}\label{eq:safe_set}
%     \mathcal{S} = \mathcal{X} \setminus \mathcal{O}.
% \end{align}
% At the beginning of the task, we assume that the robots do not have any prior information about the environment. During the control task, the robots detect obstacles and store their positions using Assumption~\ref{assump:stick_info}. At any time step $t$, let the set of obstacle constraints known to the leader (detected and stored from $t$ and during all previous time steps), and the set of obstacles detected and stored by the follower be denoted by $\mathcal{C}_{l,t}$ and $\mathcal{C}_{f,t}$, respectively. We denote:
% \begin{align*}
%     &\mathcal{C}_{l,t} \cup \mathcal{C}_{f,t} = \mathcal{O}_t,~\textnormal{with}~ \mathcal{O}_{t-1} \subseteq \mathcal{O}_t,~\forall t \leq T, 
% \end{align*}
% where $T \gg 0$ is the task duration.
%%
\section{Control Synthesis}\label{sec:control_synth}
We enumerate the steps involved in our leader-follower control synthesis briefly next, which constitute our \emph{collaborative obstacle avoidance with environment learning} algorithm. 
\begin{enumerate}[(I)]
\item At any time step $t$, the leader designs an MPC controller with horizon of $N$ steps with $NT_s\ll T$ for the joint system to reach a specified target position $S_\mathrm{tar}$, while avoiding all the stored obstacles in $\mathcal{C}_{l,t}$. This is shown in Section~\ref{ssec:lead_mpc}. 
\item If there are no obstacles in its proximity, the follower uses a control strategy to support the actions of the leader. The inference of the leader actions by the follower is described in Section~\ref{ssec:fol_inp_infer}. 
\item In the case where critical obstacles (as defined later in Definition~\ref{def:cr_d}) are detected by the follower, the follower applies an additional input contribution in order to avoid these critical obstacles, as we show in Section~\ref{ssec:fol_pol_ssec}.  
\item The leader estimates the follower's applied inputs and uses this as an ``implicit" communication to build a map of its possibly unseen obstacles lying in the follower's proximity. The leader then updates its set of known obstacles $\mathcal{C}_{l,t+T_s}$, as we show in Section~\ref{ssec:learn_obs}. The leader MPC problem is solved again at the next time step with the updated environment information.  
\end{enumerate}
We now elaborate the above steps (I)-(IV) in the following sections. The resulting \emph{collaborative obstacle avoidance with environment learning} algorithm is in Section~\ref{sec:alg}.
\subsection{Follower Policy Parameterization}\label{ssec:fol_pol_ssec}
In the set of all obstacles seen by the follower, we  define a \emph{critical obstacle point}, due to which the follower chooses to apply a reactive input.
% representation. 
%%
\begin{definition}[Critical Obstacle Points]\label{def:cr_d}
We define a critical obstacle point at time step $t$ as a \emph{point} in the set of obstacles $\mathcal{C}_{f,t}$ which is within a radius of $d_{\mathrm{cr}}$ from the follower's center of mass. Thus, the follower computes:
\begin{align} \label{eq:cr_d}
    & \mathcal{C}_{\mathrm{cr},t} = \arg \min_{c \in \mathcal{C}_{f,t}} \Vert \hat{R}^{(f)}_t - c \Vert \nonumber \\
    &~~~~~~~~~~~~~~~ \textnormal{s.t., }  \Vert \hat{R}_t^{(f)} - c \Vert  \leq d_\mathrm{cr},
\end{align}
where $\Vert \cdot \Vert$ denotes the Euclidean norm. 
\end{definition}
In case of multiple critical obstacle points satisfying \eqref{eq:cr_d}, we pick the critical obstacle point as the one that maximizes
\begin{align*}
    \max_{c \in \mathcal{C}_{\mathrm{cr},t}}\frac{\dot{\hat{R}}_t^{(f)} \cdot (c-\hat{R}_t^{(f)})}{\Vert \dot{\hat{R}}^{(f)}_t \Vert \Vert (c-\hat{R}_t^{(f)}) \Vert },
\end{align*}
that is, the one having the maximum relative velocity component towards the follower's center of mass. The inputs applied by the follower are then given by:
\begin{align}\label{eq:fol_pol_gen}
    v_t = \begin{cases} f_1(u_t),~~\textnormal{if no critical obstacle point at $t$}, \\ f_2(u_t, d_\mathrm{cr}, d_t, \phi_t)~\textnormal{otherwise},  \end{cases}
\end{align}
where $f_1(\cdot)$ and $f_2(\cdot,\cdot,\cdot,\cdot)$ can be any function chosen such that $v_t \in \mathcal{V}$, $u_t$ is the input of the leader, $d_t = \Vert \hat{R}_t^{(f)} - \mathcal{C}_{\mathrm{cr},t} \Vert$, $\phi_t$ is the angle between the vector connecting the follower center of mass to critical obstacle point and the follower center of mass to that of the leader, respectively. For our considered specific example, this is shown in Fig.~\ref{fig:sysmodel}. 
% \end{remark}

We now make the following assumption ensuring when a critical obstacle point is seen, the follower applies a separate input, as opposed to what it would have applied otherwise.
\begin{assumption}\label{assump:f_1f_2diff}
We assume in \eqref{eq:fol_pol_gen}:
\begin{align*}
    \forall t \geq 0,~\nexists~u_t, d_t, \phi_t: d_t \leq d_\mathrm{cr},~f_1(u_t) = f_2(u_t, d_\mathrm{cr}, d_t, \phi_t). 
\end{align*}
\end{assumption}
We also make the following assumption that will be used for the leader's control synthesis in Section~\ref{ssec:lead_mpc} and for learning critical obstacle points in Section~\ref{ssec:learn_obs}. 
\begin{assumption}\label{assump:lead_knows_pol}
We assume that the functions $f_1(\cdot)$ and $f_2(\cdot, \cdot, \cdot, \cdot)$ are known to the leader. 
\end{assumption}
Assumption~\ref{assump:lead_knows_pol} holds true, since such basic information can be shared offline before the task begins. Otherwise, these functions can be learned offline from data.
% We leave this as a future extension of our research. 
Our specific choice of \eqref{eq:fol_pol_gen} in this paper is given by:
\begin{align}\label{eq:fol_pol}
v_t  = \begin{cases} K_2 u_t,~\textnormal{if no critical obstacle point at $t$}, \\ K_2 u_t + K_1 (d_\mathrm{cr} - d_t) \begin{bmatrix} \cos \phi_t  \\ -\sin \phi_t  \\ 0 \end{bmatrix}~\textnormal{otherwise}, \end{cases}     
\end{align}
where in $d_t$ we directly measure $R_t$, i.e., $\hat{R}^{(f)}_t = R_t$ (see \eqref{eq:fol_states}), and the gains $K_1$ and $K_2$ known to the leader, chosen to satisfy \eqref{eq:inp_con}. Without any loss of generality in \eqref{eq:fol_pol}, we have not included a reactive torque upon seeing critical obstacle points. Hence, only the first $2 \times 2$ sub-matrix of $K_1$ need to be invertible. 
%  As incorporating such a reactive torque is trivial with slightly more computations required in Section~\ref{ssec:learn_obs}, we omit this without any loss of generality. 
% \begin{figure}[h!]
% 	\centering
% 	\includegraphics[width=0.85\columnwidth]{cr_obs.pdf}
% 	\caption{Reaction force from the chosen critical obstacle point. 
% 	}
% 	\label{fig:cr_obs}    
% \end{figure}
We choose $K_2 \in [0,1)$, and 
\begin{align*}
    K_1 = \mathrm{diag}\Big (\frac{\bar{F}_a(1-K_2)}{d_\mathrm{cr}}, \frac{\bar{F}_p(1-K_2)}{d_\mathrm{cr}}, 0 \Big ),
\end{align*}
ensuring the follower's inputs are saturated only at $d_t = 0$.
\subsection{MPC Controller of the Leader}\label{ssec:lead_mpc}
Using Assumption~\ref{assump:stick_info} and Assumption~\ref{assump:lead_knows_pol}, the constrained finite time optimal control problem that the leader has to solve for its MPC controller synthesis at time step $t$ is:
\begin{equation}\label{eq:generalized_InfOCP}
	\begin{aligned}
% V^{\star}(x_t,& \mathcal{P}_A, \mathcal{P}_B) = \notag \\
		\displaystyle\min_{U_t} &~ \displaystyle\sum\limits_{k = 1}^{N} [ (S_{t+kT_s|t} - S_{\mathrm{tar}})^\top Q_s (S_{t+kT_s|t} - S_{\mathrm{tar}}) + u_{t+(k-1)T_s|t}^\top Q_i u_{t+(k-1)T_s|t}] 
		\\
		~\text{s.t.,} & ~~ {S}_{t+kT_s|t} = f(S_{t+(k-1)T_s|t}, u_{t+(k-1)T_s|t}, v_{t+(k-1)T_s|t}), \\ % ~R_{k|t} = f_\mathcal{B}(S_{k|t}), \\ 
		&~~ \mathcal{B}(S_{t+kT_s|t}) \in \mathcal{X} \setminus \mathcal{C}_{l,t}, \\
		& ~~ u_{t+(k-1)T_s|t} \in \mathcal{U},~v_{t+(k-1)T_s|t} = f_1(u_{t+(k-1)T_s|t}),\\
% 		\text{s.t.} & ~~~ {S}_{k+1|t} = f(S_{k|t}, u_{k|t}, v_{k|t}), \\
% 		&~~~ \alpha \begin{bmatrix} s_{1,{k|t}} \\ s_{3,{k|t}} \end{bmatrix} + (1-\alpha) R_{k|t}  \in \hat{\mathcal{S}}_t,~\forall \alpha \in [0,1], \\
% 		& ~~~ R_{k|t} = \begin{bmatrix} s_{1,{k|t}} - l \cos s_{5,k|t} \\ s_{3,{k|t}} - l \sin s_{5,k|t} \end{bmatrix}, \\ 
% 		& ~~~ u_{k|t} \in \mathcal{U},~v_{k|t} = K_2 u_{k|t},\\
		&~~\forall k \in \{1,2,\dots,N\},~S_{t|t} = \hat{S}^{(l)}_t, % ~R_{t|t} = \hat{R}_t,
	\end{aligned}
\end{equation}
where $\mathcal{B}(\cdot)$ is a set of position coordinates defining the joint leader-follower system, input sequence 
% $f_\mathcal{B}(\cdot)$ is a function that relates the follower's center of mass position to the system states, 
$U_t = \{u_{t|t},\dots,u_{t+(N-1)T_s|t}\}$, $S_\mathrm{tar}$ is the target state, and $Q_s, Q_i \succcurlyeq 0$ are the weight matrices. Note, in order to avoid a mixed integer formulation arising due to all possible combinations of follower's critical obstacle points in $\mathcal{C}_{l,t}$ along the prediction horizon, in \eqref{eq:generalized_InfOCP} the leader computes the predicted $v_{k|t}$ using only $f_1(\cdot)$. 

For model \eqref{eq:mod_con} with follower policy \eqref{eq:fol_pol}, the leader uses \eqref{eq:lead_states} to estimate:
\begin{subequations}\label{eq:spec_mpc_con}
\begin{align}
    & {R}_{t+kT_s|t} = \begin{bmatrix} s_{1,{t+kT_s|t}} - (l_f + l_r) \cos s_{5,t+kT_s|t} \\ s_{3,{t+kT_s|t}} - (l_f + l_r) \sin s_{5,t+kT_s|t} \end{bmatrix}, \label{eq:con1ap}\\
    & \mathcal{B}(S_{t+kT_s|t}) = \{x: \exists \alpha \in [0,1],~ x = \alpha \begin{bmatrix} s_{1,{t+kT_s|t}} \\ s_{3,{t+kT_s|t}} \end{bmatrix} + (1-\alpha) {R}_{t+kT_s|t} \}, \\
    & v_{t+(k-1)T_s|t} = K_2 u_{t+(k-1)T_s|t} \in \mathcal{U},~\textnormal{$S_{t|t} = \hat{S}^{(l)}_{t}$}, \label{eq:con2ap}
\end{align}
\end{subequations}
in \eqref{eq:generalized_InfOCP}, for all $k \in \{1,2,\dots,N\}$, with $\mathcal{U}$ from \eqref{eq:inp_con}. Solving \eqref{eq:generalized_InfOCP}-\eqref{eq:spec_mpc_con} is difficult, mostly due to the non-convexity of the imposed state constraints $\mathcal{X} \setminus \mathcal{C}_{l,t}$, and that too for all values of parameter $\alpha \in [0,1]$. Therefore, we solve an approximation to \eqref{eq:generalized_InfOCP}-\eqref{eq:spec_mpc_con}, as shown in the Appendix. 
After finding a solution to \eqref{eq:generalized_InfOCP}, the leader applies input
\begin{align}\label{eq:mpc_pol_formulation}
    & u_t = u^\star_{t|t}
\end{align}
to joint system \eqref{eq:mod_gen} in closed-loop. Since the follower has no direct access to \eqref{eq:mpc_pol_formulation} to apply its own inputs according to \eqref{eq:fol_pol_gen}, it estimates the leader's inputs. This is elaborated next.
\subsection{Applying the Follower's Inputs}\label{ssec:fol_inp_infer}
The follower uses $\hat{S}^{(f)}_t$ to construct an estimate $\hat{u}_t$ of the leader's inputs $u_t$. This inference is done in a time duration of $\delta \ll T_s$ after time step $t$, as introduced in Fig.~\ref{fig:blk} and Section~\ref{ssec:sys_modeling}. For this inference to be feasible, we make the following sufficient assumption. 
% Let the measurements of the follower at any time step $t$ be denoted by $y_{f,t}$. 
Let $\hat{S}^{(f)}_{t} \in \mathcal{Y}_f$,~$\forall t \geq 0$.  
\begin{assumption}\label{assump:o2o}
We assume that the map from the set $\mathcal{U}$ to the set $\mathcal{Y}_f$ is invertible.
\end{assumption}
Assumption~\ref{assump:o2o} ensures that by using its set of estimates for the leader's states, the follower has the ability to uniquely infer the input $u_t$ applied by the leader.
% \textbf{this is confusing.. let's talk}
% We introduce the notation $(\cdot)_t^+$ to denote a variable's value with $\Delta t$ delay after time step $t$. 
Between the time steps $t$ and $t+\delta$ the follower applies its previous inputs $v_{t-T_s+\delta}$. Afterwards, the follower applies
\begin{align}\label{eq:fol_pol_hat}
    v_{t+\delta} = \begin{cases} f_1(\hat{u}_t),~~\textnormal{if no critical obstacle point at $t+\delta$}, \\ f_2(\hat{u}_t, d_\mathrm{cr}, d_{t+\delta}, \phi_{t+\delta}),~\textnormal{otherwise},  \end{cases}
\end{align}
where the computation of $\hat{u}_t$ uses Assumptions~\ref{assump:stick_info} and \ref{assump:o2o}. 
% Note that in \eqref{eq:fol_pol_hat} we consider the case when the follower reacts to a critical obstacle point seen at time step $t$ after a delay of $\Delta t$ (if it still remains critical). 
For our considered system model \eqref{eq:mod_con}, the follower's estimates of the joint system states (i.e., the leader's states) are given in \eqref{eq:fol_states}. 
% with evolution \eqref{eq:fol_model} considered in this paper, we use the measurements:
% \begin{equation}\label{eq:meas_fol}
%         \begin{aligned}
%         y_{f,t} = [X_{f,t}, \dot{X}_{f,t}, Y_{f,t}, \dot{Y}_{f,t}, \theta_t, \dot{\theta}_t, F_{p,t}^{(f),\mathrm{meas}}, F_{a,t}^{(f),\mathrm{meas}}]^\top,
%         \end{aligned}
% \end{equation}
% where $F_{a/p, t}^{(i), \mathrm{meas}}$ denotes the net axial/perpendicular force measured by $i \in \{l,f\}$ at time step $t$, excluding its own applied forces\footnote{can be computed by each robot as the difference between the net force measured by its on-board force sensors and its applied forces.} 
This satisfies Assumption~\ref{assump:o2o}. The construction of the estimate $\hat{u_t}$ and the corresponding form of the follower's applied inputs
\begin{align}\label{eq:v_t_final1}
    & v_{t+\delta} = \begin{cases} K_2 \hat{u}_t,~\textnormal{if no critical obstacle point at $t+\delta$}, \\ K_2 \hat{u}_t + K_1(d_\mathrm{cr} - d_{t+\delta}) \begin{bmatrix} \cos \phi_{t+\delta} \\ -\sin \phi_{t+\delta} \\ 0 \end{bmatrix}~\textnormal{otherwise}, \end{cases}
\end{align}
where $d_{t+\delta} = \Vert R_{t+\delta} - \mathcal{C}_{\mathrm{cr}, t+\delta} \Vert$, are derived in detail in the Appendix. Here we directly measure $R_{t+\delta}$, i.e., $\hat{R}^{(f)}_{t+\delta} = R_{t+\delta}$ (see \eqref{eq:fol_states}). Similar derivations can be conducted for variations of \eqref{eq:mod_con}, e.g., the rigid connections in the system replaced by elastic spring contacts, if Assumption~\ref{assump:o2o} holds. 
% %%
\subsection{Learning Critical Obstacle Points via Input Inference}\label{ssec:learn_obs}
The leader infers the reactive feedback of the follower in $v_{t+\delta}$ in \eqref{eq:fol_pol_gen} at time step $t+2\delta$. Using this, the leader's estimate of the critical obstacle point seen by the follower at time step $t+\delta$ is denoted as $\hat{\mathcal{C}}^{(l)}_{\mathrm{cr},t+\delta}$. For obtaining this estimate we first need the following assumption, along with Assumptions~\ref{assump:stick_info}-\ref{assump:lead_knows_pol} stated in Section~\ref{ssec:fol_pol_ssec}. Let $\hat{S}^{(l)}_{t} \in \mathcal{Y}_l$,~$\forall t \geq 0$. 
\begin{assumption}\label{assump:f2}
We assume that the map from the set $\mathcal{V}$ to the set $\mathcal{Y}_l$ is invertible and $f_2(\cdot, d_\mathrm{cr}, \cdot,\cdot)$ is an invertible function for any chosen value of the critical distance $d_\mathrm{cr}$.
\end{assumption}
We choose function $f_2(\cdot, d_\mathrm{cr}, \cdot, \cdot)$ satisfying Assumption~\ref{assump:f2}. Assumption~\ref{assump:f2} ensures the leader is  able to uniquely infer the critical obstacle points using its estimated follower's inputs. 
% \textcolor{red}{leader knows follower's policy?}
% Our specific choice of the follower policy parametrization \eqref{eq:fol_pol} satisfies Assumption~\ref{assump:f2}. 
Satisfying Assumptions~\ref{assump:stick_info}-\ref{assump:lead_knows_pol} and Assumption~\ref{assump:f2}, the construction of $\hat{\mathcal{C}}^{(l)}_{\mathrm{cr},t+\delta}$ for model \eqref{eq:mod_con} and follower policy \eqref{eq:fol_pol} is shown in detail in the Appendix. For this estimation the leader uses \eqref{eq:lead_states} and computes estimates of the corresponding follower states as: 
\begin{equation}\label{eq:lead_estim_fol_us}
\begin{aligned}
    & \hat{X}^{(l)}_{f,t+\delta} = X_{l,t+\delta} - (l_l + l_f) \cos \theta_{t+\delta},\\  
    & \hat{Y}^{(l)}_{f,t+\delta} = Y_{l,t+\delta} - (l_l + l_f) \sin \theta_{t+\delta},
\end{aligned}
\end{equation}
and then obtains:
\begin{align}\label{eq:cr_estim_ap1}
    \hat{\mathcal{C}}^{(l)}_{\mathrm{cr},t+\delta} = \begin{bmatrix} \hat{X}^{(l)}_{f,t+\delta} + \hat{d}_{t+\delta} \cos(\theta_{t+\delta} - \hat{\phi}_{t+\delta}) \\ \hat{Y}^{(l)}_{f,t+\delta} + \hat{d}_{t+\delta} \sin(\theta_{t+\delta} -\hat{\phi}_{t+\delta})  \end{bmatrix},
\end{align}
where $\hat{d}_{t+\delta}$ and $\hat{\phi}_{t+\delta}$ are the leader's estimate of $d_{t+\delta}$ and $\phi_{t+\delta}$, respectively. With the inferred $\hat{\mathcal{C}}^{(l)}_{\mathrm{cr},t}$, the leader then updates and uses:
\begin{align}\label{eq:C1_update}
    \mathcal{C}_{l,t+T_s} = \mathcal{C}_{l,t} \cup \delta \mathcal{C}_{l,t+T_s} \cup \hat{\mathcal{C}}^{(l)}_{\mathrm{cr},t+\delta},
\end{align}
where $\delta \mathcal{C}_{l,t+T_s}$ denotes the new obstacle constraints detected by the leader at the next time step. The process is then repeated from time step $(t\!+\!T_s)$ onward.
%%
% \begin{remark}[Inference with Unsynchronized Clocks]\label{rem:unc_inf}
% If the leader's and the follower's clocks are unsynchronized by $\Delta \leq T_s$ unknown to the leader, then the leader constructs $\hat{\mathcal{C}}_{\mathrm{cr},t+\delta}$ using its estimates $\hat{S}^{(l)}_{t+T_s}$ at the end of its time step, instead of using $\hat{S}^{(l)}_{t+2\delta}$. The output of the inference remains unchanged. 
% \end{remark}
%%
% \begin{remark}\label{rem:alt_cr}
% One can choose any alternative representation of a critical obstacle as opposed to a point representation such as in \eqref{eq:cr_d}, as long as the Requirement on $f_2(\cdot,\cdot,\cdot,\cdot)$, and Assumption~\ref{assump:f_1f_2diff}-\ref{assump:lead_knows_pol} hold. That is, from the reactive inputs of the follower, as unique critical obstacle point can be inferred by the leader. We have nonetheless chosen such a point representation, as this avoids any possible over-approximation of the obstacles in $\mathcal{C}_{f,t}$. 
% \end{remark}
%%
\subsection{Leader-Follower Role Switching}\label{switch_ssec}
% In our proposed method, the leader infers its unknown environment constraints using implicit communications from the follower. 
Although the leader learns $\hat{\mathcal{C}}^{(l)}_{\mathrm{cr},t+\delta}$ and updates its controller, this can still lead to failure in avoiding obstacles in tight environments. For example, if the follower approaches a tight corner with multiple obstacles, the leader may not have sufficient time to  generate a feasible trajectory for the joint system, as it does not directly detect the whole obstacle map from the follower and infers \emph{only} the critical obstacle points detected by the follower. Therefore, switching the roles of the leader and the follower in these scenarios can be useful, enabling the leader to directly see all the obstacles in the tight corner. Such a role switching strategy of the leader and the follower is motivated by \cite{losey2020learning}, where the roles are switched with a fixed frequency. In general, we define a time dependent role switching function for an agent as:
\begin{align}\label{eq:swt_gen}
    f_\mathrm{swt}: (x, \mathcal{C}, t) \mapsto \{0,1\}, 
\end{align}
where $x$ and $\mathcal{C}$ denote the switching deciding states and obstacles of the agent, respectively, and 0 denotes no switching and 1 denotes a switch trigger. 

% In the next section, we pick a specific time invariant switching function for our considered example with model \eqref{eq:mod_con} and then present the corresponding algorithm summarizing Sections~\ref{ssec:fol_pol_ssec}-\ref{ssec:learn_obs}. 
%%
\subsection{Algorithm}\label{sec:alg}
We summarize our proposed collaborative obstacle avoidance with environment learning algorithm with system model \eqref{eq:mod_con} and follower policy parametrization \eqref{eq:fol_pol} in Algorithm~\ref{alg1}. 
\begin{algorithm}[h]
    \caption{
    Collaborative Obstacle Avoidance with Environment Learning  
    }
    \label{alg1}
    \begin{algorithmic}[1]
      \Statex \hspace{-1.2em}\textbf{Initialize:} $t = 0$, $v_{(t-T_s)+\delta}, S_0$ 
      
      \Statex \hspace{-1.2em}\textbf{Inputs:} $S_\mathrm{tar}$, $Q_i$, $Q_s$, $d_\mathrm{cr}$, $\mathcal{U}$, $K_1, K_2, T, \delta, N, T_s, \mathcal{X}$  
      
      \Statex \hspace{-1.2em}\textbf{Data:} $\mathcal{C}_{l,t}$,  $\mathcal{C}_{f,t}$
       
      \vspace{1.2mm}
      
      \WHILE{$t \leq T$}
      \STATE \textit{Leader at $t$}: Get $u_t$ from \eqref{eq:generalized_InfOCP}-\eqref{eq:mpc_pol_formulation}. Apply to~\eqref{eq:mod_con};
      
      \STATE \textit{Follower at $t$}: Apply $v_{(t-T_s)+\delta}$ to \eqref{eq:mod_con} in $[t,t+\delta)$;

      \STATE \textit{Follower at $t+\delta$}: Compute $\mathcal{C}_{\mathrm{cr},t+\delta}$ with \eqref{eq:cr_d};

      \Statex \hspace{96pt} Obtain $\hat{R}^{(f)}_{t+\delta}$ from \eqref{eq:fol_states}; 
      
      \Statex \hspace{96pt} Infer $\hat{u}_t$ (see Appendix);
      
     \Statex \hspace{96pt} Apply $v_{t+\delta}$ in \eqref{eq:v_t_final1} to \eqref{eq:mod_con};
      
      \STATE \textit{Leader at $t+2\delta$}:  Obtain $\hat{R}^{(l)}_{t+\delta}$ from \eqref{eq:lead_estim_fol_us}; 
      
      \Statex \hspace{92pt}  Obtain $\hat{S}^{(l)}_{t+2\delta}$ from \eqref{eq:lead_states};
      
      \Statex \hspace{92pt} Estimate $\hat{v}_{t+\delta}$. Get $\hat{d}_{t+\delta}$, $\hat{\phi}_{t+\delta}$ from \eqref{eq:v_t_final1}; 
      
      \Statex \hspace{92pt} Estimate $\hat{\mathcal{C}}^{(l)}_{\mathrm{cr}, t+\delta}$ with \eqref{eq:cr_estim_ap1}; 
      
     \STATE \textit{Follower at $t+T_s$}: Check $f_\mathrm{swt}(\hat{R}^{(f)}_{t+\delta}, \mathcal{C}_{\mathrm{cr},t+\delta})$ and pick switch;
     
     \STATE \textit{Leader at $t+T_s$}: Check $f_\mathrm{swt}(\hat{R}^{(l)}_{t+\delta}, \hat{\mathcal{C}}^{(l)}_{\mathrm{cr},t+\delta})$ and pick switch;
     
      \STATE $t = t+T_s$; 
      
      \ENDWHILE
      \end{algorithmic}
\end{algorithm}
As a specific choice for \eqref{eq:swt_gen}, we pick:
\begin{align}\label{eq:role_switch_strt}
    f_\mathrm{swt}(R_{t+\delta}, \mathcal{C}_{t+\delta}) = \begin{cases}  1,~\textnormal{if $\Vert R_{t+\delta} - \mathcal{C}_{t+\delta} \Vert \leq d_\mathrm{thr}$}, \\ 0~\textnormal{otherwise}, \end{cases}
\end{align}
where $d_\mathrm{thr}$ is a chosen distance threshold value, and the follower and the leader use $\mathcal{C}_{t+\delta} = \mathcal{C}_{\mathrm{cr},t+\delta}$, $\mathcal{C}_{t+\delta} = \hat{\mathcal{C}}^{(l)}_{\mathrm{cr},t+\delta}$, and $R_{t+\delta} = \hat{R}^{(f)}_{t+\delta}$ and $R_{t+\delta} = \hat{R}^{(l)}_{t+\delta}$ obtained from \eqref{eq:estimates_us}, respectively. Having evaluated \eqref{eq:role_switch_strt} at time step $t+\delta$, the agents decide the role switch trigger accordingly for control design at $t+T_s$. Since the error between $\hat{R}^{(f)}_{t+\delta}$ and $\hat{R}^{(l)}_{t+\delta}$ is zero (see \eqref{eq:estimates_us}), the switch happens simultaneously at $t+T_s$ without any explicit communication, if the leader has an accurate estimate~\eqref{eq:cr_estim_ap1}. We alternately keep changing the cost in \eqref{eq:generalized_InfOCP}-\eqref{eq:spec_mpc_con} with role switches, always penalizing the deviation of the initial leader from $S_\mathrm{tar}$.

\begin{remark}[Unsynchronized Clocks]
If the leader's and the follower's clocks are unsynchronized, the order of operations shown in Algorithm~\ref{alg1} can no longer be ensured. However, for a small sampling time $T_s$, the performance of Algorithm~\ref{alg1} does not change noticeably. This was observed during the numerical experiments in Section~\ref{sec:numerics} with the chosen value of $T_s$ in Table~\ref{tab1}. In such cases, check \eqref{eq:role_switch_strt} may 
% when the switch triggers from don't synchronize, 
% by $\Delta \leq T_s$ unknown to each robot, then both switch separately. The follower switches first into a leader, and then begins its inference of critical obstacles (Section~\ref{ssec:learn_obs}, Remark~\ref{rem:unc_inf}) after waiting for another time step. This ensures the derivation of \eqref{eq:cr_estim_ap1} stays valid (see Appendix), at the expense of missing only one inference point over that $T_s$ duration. For time $\Delta$, 
result in two leaders for a fraction of the sampling time, when both robots apply MPC controllers by solving \eqref{eq:generalized_InfOCP}-\eqref{eq:mpc_pol_formulation}.
% , when the leader roles overlap.   
\end{remark}

\begin{remark}[Imperfect Estimation]
If the estimate  $\mathcal{C}^{(l)}_{\mathrm{cr},t+\delta}$ has large errors, one may alternatively decide role switches with a fixed time period, similar to~\cite{losey2020learning}. In such a case, the leader may not include $\mathcal{C}^{(l)}_{\mathrm{cr},t+\delta}$ in \eqref{eq:C1_update}, and instead include a time varying penalty in \eqref{eq:generalized_InfOCP} based on its inferred obstacles.  
\end{remark}
\section{Numerical Experiments}\label{sec:numerics}
We present our numerical simulations in this section. First we detail the problem setup, and then we compare the results from Algorithm~\ref{alg1} with two alternative strategies. We consider synchronous clocks for these experiments. The source codes are in: \href{https://github.com/monimoyb/LeadFollowRobots}{\textit{github.com/monimoyb/LeadFollowRobots}}. We use Python 3.7.3 and the SLSQP solver in SciPy 1.6.

\subsection{Experimental Setup}
% We start with the shown system and environment configuration in Fig.~\ref{fig:fig1}. 
The parameters of the problem are shown in Table~\ref{tab1}. 
\begin{table}[h]
\centering
\caption{Table of parameter values. Note, the results presented are after relaxing $\delta \ll T_s$.}
\label{tab1}
\begin{tabular}{ |c| c| c |c| } 
 \hline
 Parameter & Value & Parameter & Value \\
 \hline
 $m_l$ & 0.04 kg & $m_f$ & 0.04 kg \\
 $l_l$ & 0.8 m & $l_f$ & 0.8 m \\ 
 $m_r$ & 0.01 kg & $T_s$ & 0.03 s \\ 
%  $S_\mathrm{tar}$ & $[3,0,3.95,0,0,0]^\top$ & $S_0$ & $[7.5, 0, 7.2, 0, 0.1, 0]^\top$ \\ % 
 $d_\mathrm{cr}$ & 1.4m & $K_2$ & 0.5\\
 $\bar{F}_a$ & 5 N & $\bar{F}_p$ & 5 N\\
 $\bar{\tau}$ & 0.5 Nm & $\delta$ & 0.02 s\\
 $N$ & 3 & $T$ & 2.7 s \\
 \hline
\end{tabular}
% \vspace{-10pt}
\end{table}
We consider the set of obstacles in  $\mathcal{C}_{l,t}$ and $\mathcal{C}_{f,t}$ as a collection of discrete point coordinates, since we simulate the detection of these obstacles with a lidar-like angle sweep. Both the leader and the follower record point cloud information of surrounding obstacles lying within a radius of $1.2$ meters, with a resolution of $3.6$ degrees. The critical distance, as defined in \eqref{eq:cr_d}, is chosen as $d_\mathrm{cr} = 1.1$ meters. 
 
We solve the leader's MPC controller synthesis problem (see Appendix for this approximation to \eqref{eq:generalized_InfOCP}-\eqref{eq:spec_mpc_con}) with semi-definite weight matrices  $Q_s = \mathrm{diag}(120,4,120,4,0,0.01)$, and $Q_i = \mathrm{diag}(0.05, 0.05, 0.01)$ and with  $S_\mathrm{tar} = [3,0,3.95,0,0,0]^\top$ and $S_0 = [7.5, 0, 7.2, 0, 0.1, 0]^\top$. 
% We use Python 3.7.3 and the SLSQP solver in SciPy 1.5.3.
% \begin{figure}[h!]
% 	\centering
% 	\includegraphics[width=\columnwidth]{obs_fig.eps} %\\
% 	\caption{Layout of the obstacles in the room along with the initial configuration of the system. The red dot represents the leader and the blue dot represents the follower. The red star denotes the target position of the leader.
% 	}
% 	\label{fig:fig1}    
% \end{figure}
%%
\subsection{Trajectory Comparison with Alternative Strategies}\label{ssec:role_num}
We now present the results of two alternative strategies and compare them with the ones from Algorithm~\ref{alg1}. In all the following figures, the red dot represents the leader and the blue dot represents the follower. The red star denotes the target position of the leader.
\subsubsection{No Environment Learning}
\begin{figure}[h!]
	\centering
	\includegraphics[width=0.6\columnwidth]{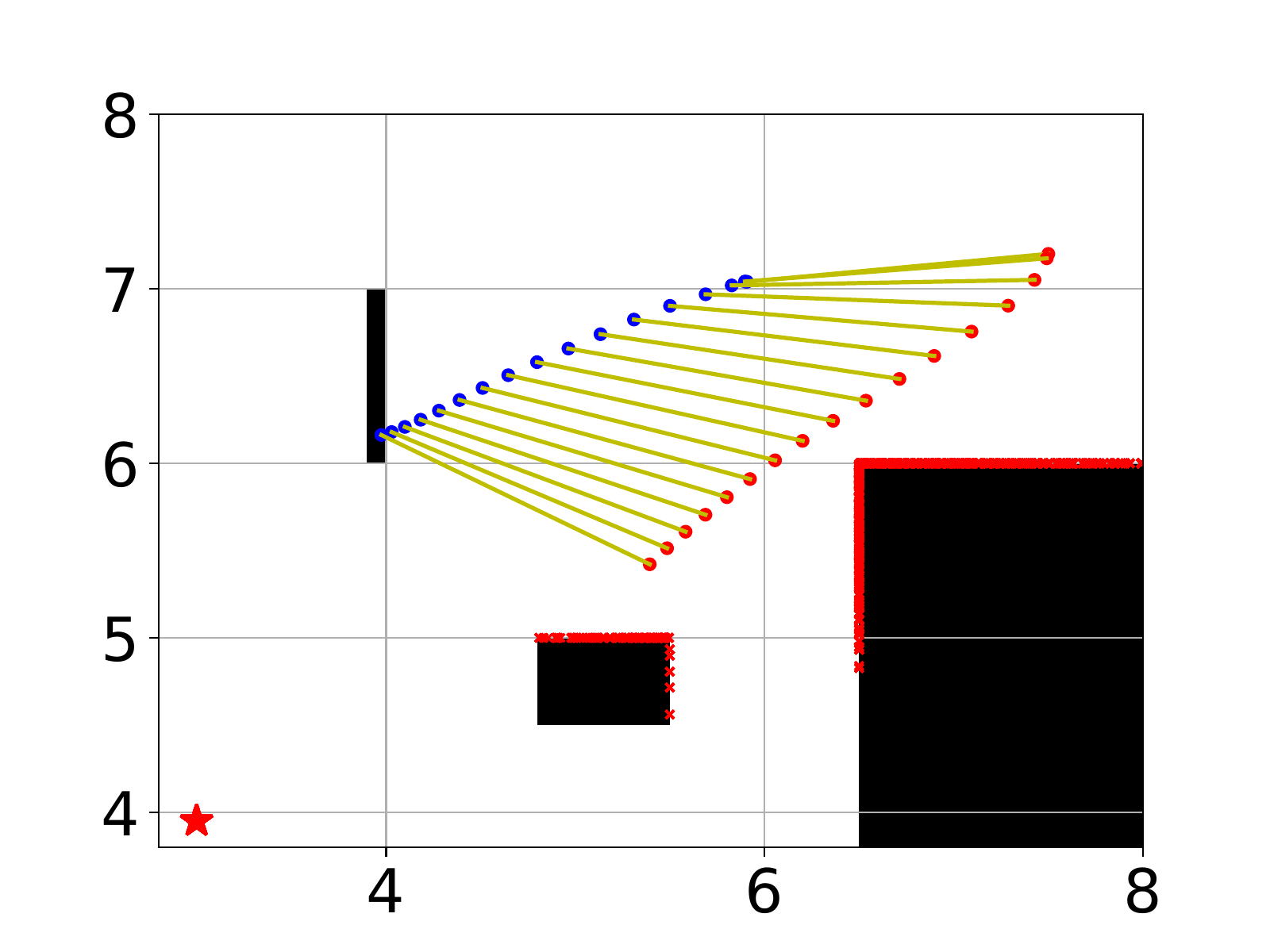}%\\
	\caption{Trajectory snapshot of the rod without learning obstacles from the follower inputs. The task fails. 
	}
	\label{fig:firstfig}    
\end{figure}

The first strategy is an MPC based standard leader-follower obstacle avoidance strategy motivated by \cite{li2019dynamical,desai1999motion,bharatheesha2017dynamic,swag,shorinwa2020scalable}, where the leader applies the MPC controller \eqref{eq:generalized_InfOCP}-\eqref{eq:mpc_pol_formulation}, the follower applies \eqref{eq:fol_pol}, and the leader \emph{does not infer} any obstacle information from the follower's inputs. So the set of obstacles used by the leader for MPC design is updated as:
\begin{align*}
    \mathcal{C}_{l,t+T_s} = \mathcal{C}_{l,t} \cup \delta \mathcal{C}_{l,t+T_s},~\forall t \geq 0.
\end{align*}
The trajectory of the joint system with this strategy is shown in Fig.~\ref{fig:firstfig}.
The red crosses denote the obstacle points directly seen by the leader, which are successfully avoided. However, the rod hits the left most obstacle around position (4,6.2) on the follower's side. This obstacle remains unknown to the leader, as it is not inferred from follower inputs.
\subsubsection{No Role Switching} 
\begin{figure}[h!]
	\centering
	\includegraphics[width=0.6\columnwidth]{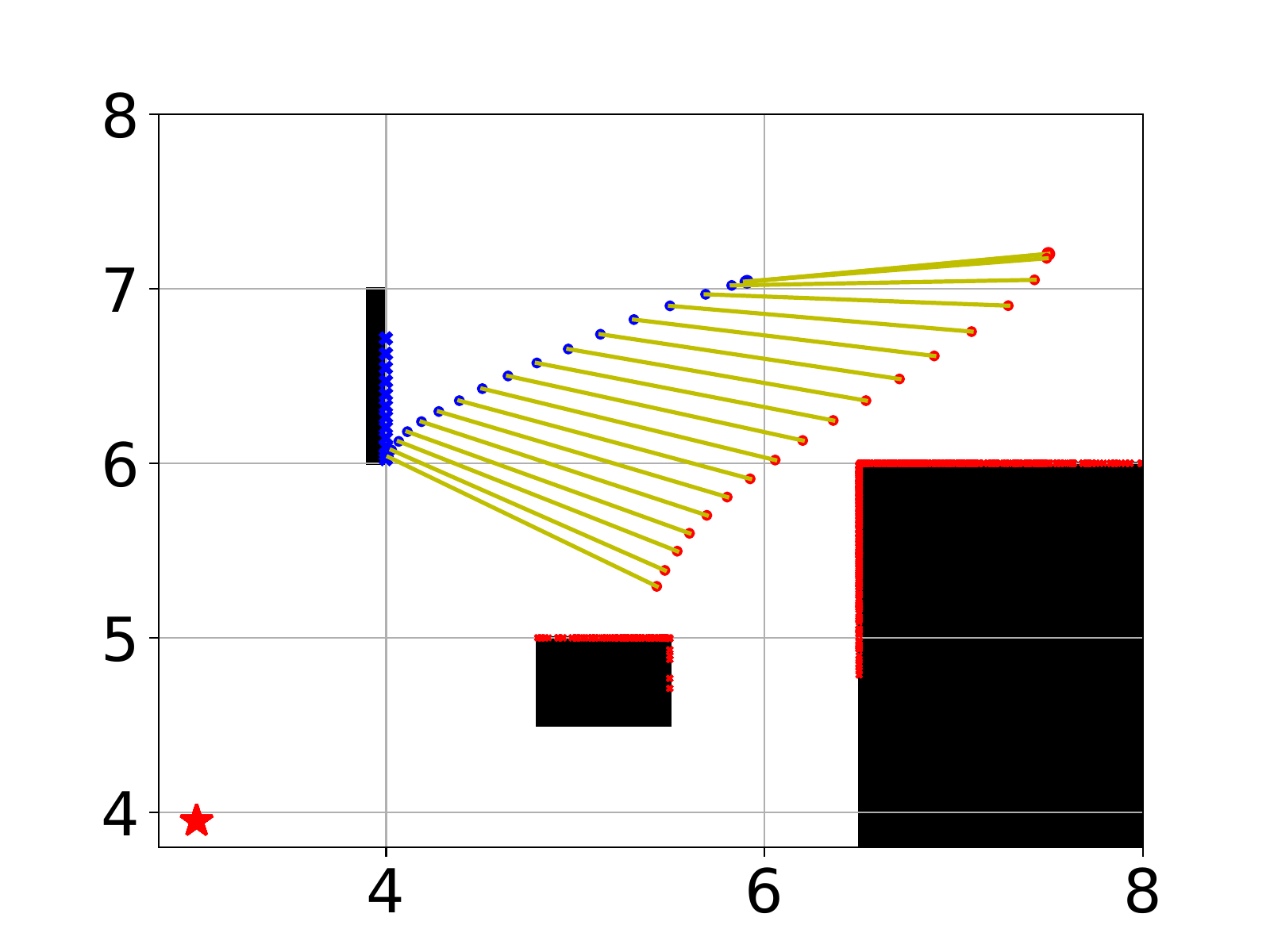}%\\
	\caption{Trajectory snapshot of the rod with a fixed leader-follower role allotment and learning obstacles from the follower inputs. The task still fails. 
	}
	\label{fig:noswitch}    
\end{figure}

The second strategy is similar to Algorithm~\ref{alg1}, with the exception that there is \emph{no switching} of the leader-follower roles. Such a fixed role assignment is a standard practice in the literature, e.g., \cite{stilwell1993toward, gross2006transport,  wang2016force,swag}. 
% That is,
% \begin{align*}
%     f_\mathrm{swt}(R_{t+\delta}, \mathcal{C}_{t+\delta}) = 0,~\forall t \geq 0,
% \end{align*}
% with $\mathcal{C}_{t+\delta} = \hat{\mathcal{C}}_{\mathrm{cr},t+\delta}^{(l)}$, $\mathcal{C}_{t+\delta}=\mathcal{C}_{\mathrm{cr},t+\delta}$, and $R_{t+\delta} = \hat{R}^{(l)}_{t+\delta}$ and $R_{t+\delta} = \hat{R}^{(f)}_{t+\delta}$ obtained from \eqref{eq:estimates_us} for the leader and the follower, respectively. 
As opposed to Strategy 1, here we update the leader's known set of obstacles as \eqref{eq:C1_update}, having inferred critical obstacle points \eqref{eq:cr_estim_ap1} using the follower's estimated inputs. The trajectory of the joint system with this strategy is shown in Fig.~\ref{fig:noswitch}. The blue crosses denote the follower's critical obstacle points inferred by the leader.
As seen in Fig.~\ref{fig:noswitch}, the follower still collides with the left most obstacle around position (4,6), despite the leader learning additional blue obstacle points using the follower's feedback. This shows that a fixed role allotment here is inhibiting.    
\subsubsection{Algorithm~\ref{alg1}}
\begin{figure}[h!]
	\centering
	\includegraphics[width=0.6\columnwidth]{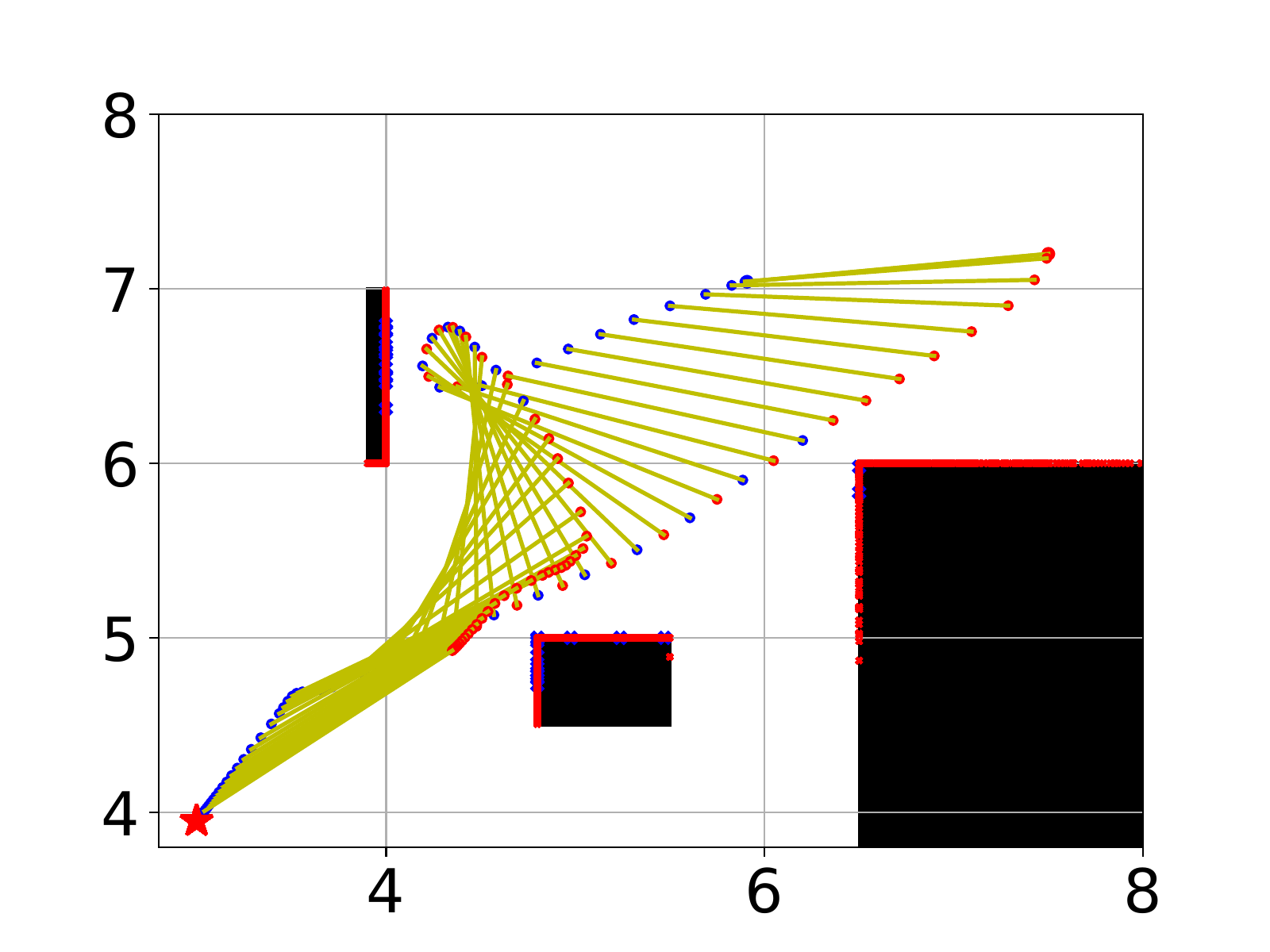}%\\
	\caption{Trajectory snapshot of the rod with a switching leader-follower role allotment and learning obstacles from the follower inputs. The task succeeds.}
	\label{fig:switch}    
\end{figure}

We now demonstrate the results using Algorithm~\ref{alg1}, where we switch the roles of the leader and the follower using \eqref{eq:role_switch_strt} and a threshold distance of $d_\mathrm{thr} = 0.8$ meters. The trajectory of the joint system is shown in Fig.~\ref{fig:switch}. 
As seen in Fig.~\ref{fig:switch}, the joint system now successfully avoids all the obstacles after incorporating the leader-follower switch. After the switch, the leader directly faces the left most obstacle and collects multiple cloud points on its surface (the red crosses). These additional obstacle cloud points are missing in Fig.~\ref{fig:firstfig}, where there is no obstacle inference by the leader, and also in Fig.~\ref{fig:noswitch}, where the leader relies on the follower for inferring only one critical obstacle point (the blue crosses) at a time. The task also succeeds, as the initially chosen leader reaches $S_\mathrm{tar}$ by $T=2.7$ seconds.
% Furthermore, the initially chosen leader robot reaches the target position (as a follower after multiple switches), thus successfully completing the transportation task. 

\subsection{Multiple Trials with Varying Environment}
\begin{figure}[h!]
\vspace{-10pt}
	\centering
	\includegraphics[width=0.6\columnwidth]{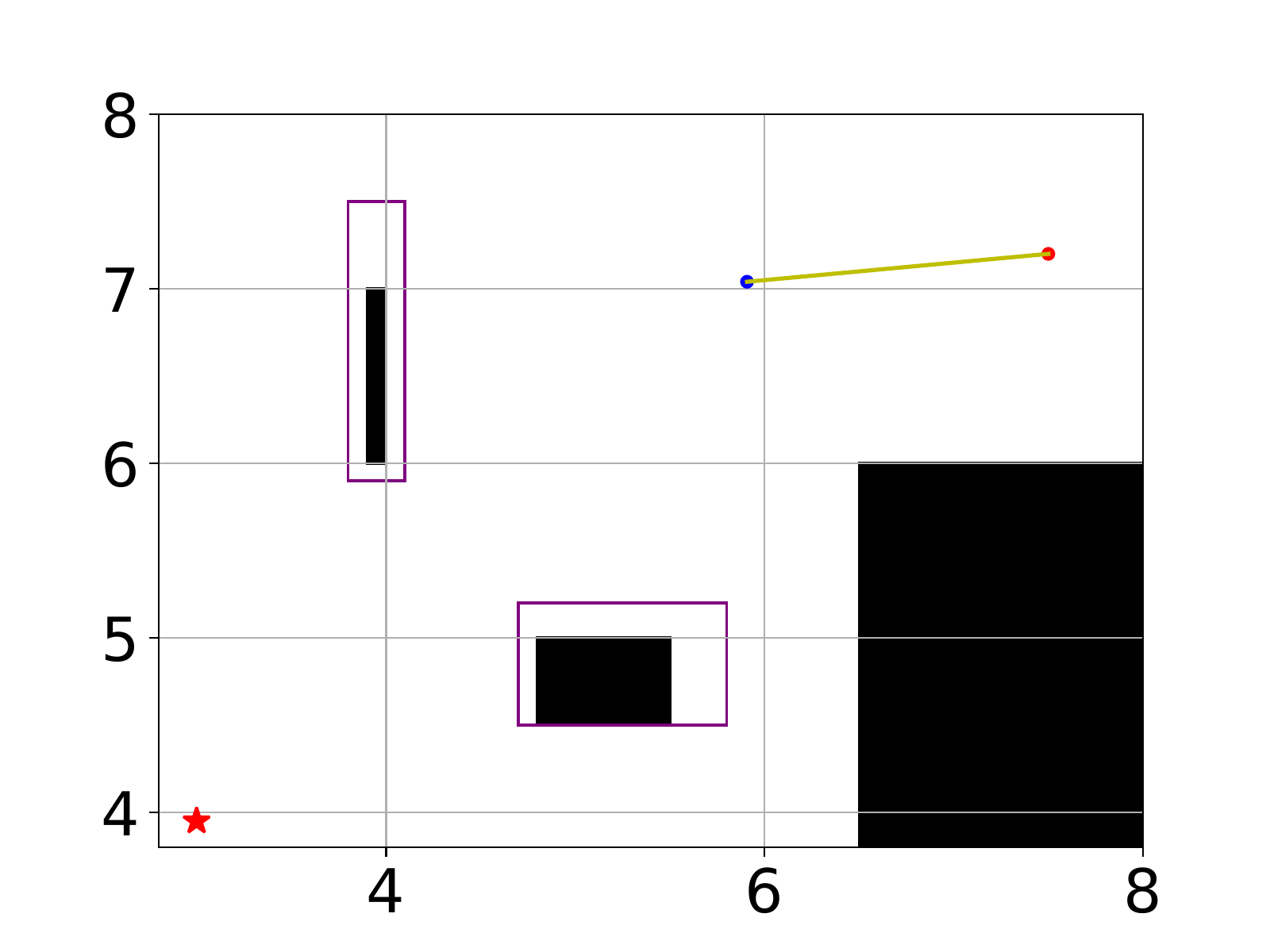}%\\
	\caption{Zones containing varying obstacle positions with the given joint system's initial configuration.}
	\label{fig:rand_env}    
\end{figure}

We now conduct 100 trials with each of the above three strategies, with varying positions of the left most obstacle and the one at the center of the environment. The variations are contained in the purple regions shown in Fig.~\ref{fig:rand_env}. The shape and the size of the obstacles are unchanged, and one of their vertices is chosen uniformly in the shown regions.
Successful trials are only recorded if the joint system avoids all the obstacles and the initially chosen leader robot reaches a neighborhood of radius 0.5 meters around $S_\mathrm{tar}$ within $T=2.7$s, i.e., 90 steps. Table~\ref{table:results} summarizes the results.
\begin{table}[h]
\centering
\caption{Strategy comparison across 100 trials. Strategy 1: No Environment Learning, Strategy 2: No Role Switching, Strategy 3: Algorithm~\ref{alg1}. CFT denotes: Collision Free Trials. }
\label{table:results}
\begin{tabular}{ |c|c|c|c|} 
 \hline
 Feature & Strategy 1 & Strategy 2 & Strategy 3 \\
 \hline
 Successful Trials ($\%$) & 0 & 24 & 86 \\
 Collision Failures ($\%$) & 100 & 56 & 14 \\ 
 Timed-Out Failures ($\%$) & 0 & 20 & 0 \\
 Avg. $\#$ of Steps in a CFT & N/A & 87 & 49\\
 \hline
\end{tabular}
% \vspace{-18pt}
\end{table}
Table~\ref{table:results} shows that Algorithm~\ref{alg1} outperforms the rest with the highest success rate, while reaching the target neighborhood fastest. 
% \begin{itemize}
%     \item With No Environment Learning:~$0\%$,
%     \item With No Role Switching:~$59\%$,
%     \item With Algorithm~\ref{alg1}:~$75\%$.
% \end{itemize}
% \textbf{depending how much time we have wee need to generate a table with distributions.. saying while experimental test i s objective of future research in the videos...}
%%
\section{Conclusion}
We proposed a leader-follower strategy for a two-robots collaborative transportation task in a partially known environment with obstacles. 
% We consider no  explicit communication between the robots. %
The leader solves an MPC problem at any given time with its known set of obstacles to plan a feasible trajectory and complete the task. The follower's policy is designed to assist the leader, but also react to additional obstacles in proximity which might be unseen to the leader. The difference between the predicted and the actual follower inputs is used by the leader to infer additional unseen environment constraints. 
% The set of known constraints for the MPC problem is then updated at the next step. 
We also propose a switching strategy for the leader-follower roles,
% which requires no explicit communication, 
improving the control performance in tight environments. Our algorithm outperforms two alternative strategies in the demonstrated numerical example, with the lowest collision rate and the fastest average task completion speed. 
%%%%
\section*{Acknowledgements}
This project is funded by grants ONR-N00014-18-1-2833, and NSF-1931853. The project has also received funding from the European Union’s Horizon 2020 research and innovation programme under the Marie Sk\l{}odowska-Curie grant agreement No 846421.
%%
% \newpage 
\renewcommand{\baselinestretch}{1}
\bibliographystyle{IEEEtran}
\bibliography{bibliography.bib}
\newpage
\section*{Appendix}
\subsection{Tractable Approximation of \eqref{eq:generalized_InfOCP}-\eqref{eq:spec_mpc_con}}
We first introduce the following notation: let $\mathrm{dist}(x,\mathcal{C})$ denote the shortest distance in $\ell_2$ norm from point $x$ to all points in the set $\mathcal{C}$. Then the approximation of \eqref{eq:generalized_InfOCP}-\eqref{eq:spec_mpc_con} is:
\begin{equation}\label{eq:opt_appendix}
	\begin{aligned}
% V^{\star}(x_t,& \mathcal{P}_A, \mathcal{P}_B) = \notag \\
		\displaystyle\min_{U_t} &~~ \displaystyle\sum\limits_{k = 1}^{N} \Bigg [ (S_{t+kT_s|t} - S_{\mathrm{tar}})^\top Q_s (S_{t+kT_s|t} - S_{\mathrm{tar}}) + \sum_{i=1}^{n} \ell \Big ( \mathrm{dist} (\alpha_i \begin{bmatrix} s_{1,{t+kT_s|t}} \\ s_{3,{t+kT_s|t}} \end{bmatrix} + (1-\alpha_i) R_{t+kT_s|t}, \mathcal{C}_{l,t}) \Big )+ \\
		&~~~~~~~~ \cdots + u_{t+(k-1)T_s|t}^\top Q_i u_{t+(k-1)T_s|t} \Bigg ]
		\\
		\text{s.t.,} & ~~~ {S}_{t+kT_s|t} = f(S_{t+(k-1)T_s|t}, u_{t+(k-1)T_s|t}, v_{t+(k-1)T_s|t}),\\
		&~~~\eqref{eq:con1ap}~\textnormal{and } \eqref{eq:con2ap},~\forall k \in \{1,2,\dots,N\}, 
% 		&~~~S_{t|t} = \hat{S}^{(l)}_t,~\textnormal{from \eqref{eq:lead_states}},
	\end{aligned}
\end{equation}
where $\ell(\cdot)$ is a positive definite cost function and $\alpha_i \in [0,1]$ for $i \in \{1,2,\dots, n\}$ are $n$ sampled values of parameter $\alpha$.
% ($\alpha_i = 0,1$ included). 
% Note, we include $\alpha_i = 0,1$ corresponding to the leader and the follower positions, respectively, in these $n$ samples. 

% To summarize, we have removed the state constraints $\hat{\mathcal{S}}_t$ from \eqref{eq:generalized_InfOCP}, and have included a penalty $\ell(\cdot)$ in the cost function in \eqref{eq:opt_appendix} which penalizes the proximity of the rod to any known obstacle. Although this guarantees the feasibility of \eqref{eq:opt_appendix} at all time steps $t \geq 0$, the rod may now violate the safety constraint $\hat{\mathcal{S}}_t$ and hit obstacles. In that case, we re-tune the weights $Q_s, Q_i$ and the cost function $\ell(\cdot)$. 
%%
\subsection{Applying the Follower's Inputs for Model \eqref{eq:mod_con}}
Recall the follower estimates from \eqref{eq:fol_states}. As mentioned in Section~\ref{ssec:fol_inp_infer}, the follower obtains these estimates at $t+\delta$, denoted by $\dot{\hat{X}}_{l,t+\delta}^{(f)}, \dot{\hat{Y}}_{l,t+\delta}^{(f)}$ and $\dot{\theta}_{t+\delta}$. The follower then uses the following equations: 
      \begin{equation*}
        \begin{aligned}
         & \dot{\hat{X}}_{l,t+\delta}^{(f)} = \dot{\hat{X}}^{(f)}_{l,t} + \hat{q}_1 \delta,~\dot{\hat{Y}}_{l,t+\delta}^{(f)} = \dot{\hat{Y}}^{(f)}_{l,t} + \hat{q}_2 \delta,~\textnormal{and }\dot{\theta}_{t+\delta} = \dot{\theta}_t + \hat{\mathcal{T}} \delta,~\textnormal{with}\\
         & \hat{\mathcal{T}} = \frac{(-F_{pf, t-T_s+\delta}l_f + \hat{F}^{(f)}_{pl, t}l_l + \tau_{f,t-T_s+\delta} + \hat{\tau}^{(f)}_{l,t})}{J},\\
         & \hat{q}_1  = -(l_l \sin \theta_t \hat{\mathcal{T}} + l_l \cos \theta_t \dot{\theta}_t^2) + \frac{1}{m} (\cos \theta_t(\hat{F}^{(f)}_{al, t} + F_{af, t-T_s+\delta}) - \sin \theta_t (\hat{F}^{(f)}_{pl,t} + F_{pf,t-T_s+\delta})),\\
         &     \hat{q}_2  = (l_l \cos \theta_t \hat{\mathcal{T}} - l_l \sin \theta_t \dot{\theta}_t^2) + \frac{1}{m} (\sin \theta_t (\hat{F}^{(f)}_{al,t} + F_{af,t-T_s+\delta})) + \cos \theta_t (\hat{F}^{(f)}_{pl,t} + F_{pf,t-T_s+\delta})), 
        \end{aligned}
         \end{equation*}
%\fi
solves for $\hat{u}_t = [\hat{F}^{(f)}_{al,t}, \hat{F}^{(f)}_{pl,t}, \hat{\tau}^{(f)}_{l,t}]^\top$, and applies \eqref{eq:v_t_final1}. Note, Assumption~\ref{assump:o2o} is satisfied, as the leader's inputs appear in the above set of equations linearly and have unique solutions. 
\renewcommand{\baselinestretch}{0.8}
\subsection{Learn Critical Obstacles with Model \eqref{eq:mod_con} and Policy \eqref{eq:fol_pol}}
At time step $t+2\delta$, just after the follower applies \eqref{eq:v_t_final1}, the leader has access to its state estimates (i.e., directly measured) using \eqref{eq:lead_states}. It then uses the following equations:
      \begin{equation*}
        \begin{aligned}
         & \dot{{X}}_{l,t+2\delta} = \dot{{X}}_{l,t+\delta} + \hat{q}_1 \delta,~\dot{{Y}}_{l,t+2\delta} = \dot{{Y}}_{l,t+\delta} + \hat{q}_2 \delta,~\textnormal{and }\dot{\theta}_{t+2\delta} = \dot{\theta}_{t+\delta} + \hat{\mathcal{T}} \delta,~\textnormal{with}\\
         & \hat{\mathcal{T}} = \frac{(-\hat{F}^{(l)}_{pf, t+\delta}l_f + {F}_{pl, t}l_l + \hat{\tau}^{(l)}_{f,t+\delta} + {\tau}_{l,t})}{J},\hat{\tau}^{(l)}_{f,t+\delta} = K_2\tau_{l,t},\\
         & \hat{q}_1  = -(l_l \sin \theta_{t+\delta} \hat{\mathcal{T}}+ l_l \cos \theta_{t+\delta} (\dot{\theta}_{t+\delta})^2) + \frac{1}{m} (\cos \theta_{t+\delta} \times ({F}_{al, t}+ \hat{F}^{(l)}_{af, t+\delta}) - \sin \theta_{t+\delta} ({F}_{pl,t} + \hat{F}^{(l)}_{pf,t+\delta})),\\
         &     \hat{q}_2  = (l_l \cos \theta_{t+\delta} \hat{\mathcal{T}}- l_l \sin \theta_{t+\delta} (\dot{\theta}_{t+\delta})^2) + \frac{1}{m} (\sin \theta_{t+\delta} \times ({F}_{al,t} + \hat{F}^{(l)}_{af,t+\delta})) + \cos \theta_{t+\delta} ({F}_{pl,t} + \hat{F}^{(l)}_{pf,t+\delta})), 
        \end{aligned}
         \end{equation*}
and solves for $\hat{v}_{t+\delta} = [\hat{F}^{(l)}_{af,t+\delta}, \hat{F}^{(l)}_{pf,t+\delta}, \hat{\tau}^{(l)}_{f,t+\delta} ]^\top$. Note, Assumption~\ref{assump:f2} is satisfied, as the follower's inputs appear in the above set of equations linearly and have unique solutions. The leader infers $d_{t+\delta}$ and $\phi_{t+\delta}$ by solving:
\begin{subequations} \label{eq:extra_con}
\begin{align} 
  & \hat{\phi}_{t+\delta} = -\arctan \Bigg (\frac{\hat{F}^{(l)}_{pf,t+\delta} - K_2 F_{pl,t}}{\hat{F}^{(l)}_{af,t+\delta}-K_2F_{al,t}} \Bigg ), \\
  & \Vert \hat{F}^{(l)}_{af,t+\delta}-K_2F_{al,t} \Vert_2^2 = \Bigg ( \frac{\bar{F}^2_a(1-K_2)^2}{d^2_\mathrm{cr}} \cos^2 \hat{\phi}_{t+\delta} + \frac{\bar{F}^2_p(1-K_2)^2}{d^2_\mathrm{cr}} \sin^2 \hat{\phi}_{t+\delta} \Bigg ) \times (d_\mathrm{cr}-\hat{d}_{t+\delta})^2,
\end{align}
\end{subequations}
where $\bar{F}_{a/p}$ is the axial/perpendicular force constraint, as shown in \eqref{eq:inp_con}. The leader then uses \eqref{eq:extra_con} in \eqref{eq:lead_estim_fol_us} to infer the critical obstacle point using \eqref{eq:cr_estim_ap1}.

\end{document}